\documentclass{article}
\usepackage{makeidx}  
\usepackage{epsfig} 
\usepackage{url}
\usepackage[latin1]{inputenc}
\usepackage{color}
\usepackage{alltt}

\begin{document} 

\title{Improved evolutionary generation of XSLT stylesheets\thanks{Supported by projects TIN2007-68083-C02-01, and P06-TIC-02025.}} 
\author{Pablo Garc\'ia-S\'anchez, JLJ Laredo, JP Sevilla, Pedro Castillo, JJ Merelo\\
Depto. Arquitectura y Tecnolog\'ia de Computadoras\\
ETSIIT - Universidad de Granada (Spain)\\
E-Mail: \texttt{jj@merelo.net}}

\maketitle              

\begin{abstract} 
This paper introduces a procedure based on genetic programming to evolve
XSLT programs (usually called stylesheets or logicsheets). XSLT is a
general purpose, document-oriented functional language, generally used
to transform XML documents (or, in general, solve any problem that can
be coded as an XML document). The
proposed solution uses a tree representation for the stylesheets as
well as diverse specific operators in order to obtain, in the studied
cases and a reasonable time, a XSLT stylesheet that performs the
transformation. Several types of representation have been compared,
 resulting in different performance and degree of success.

\end{abstract} 

\section{Introduction}

XML (eXtensible Markup Language, \cite{xml:bible,LearningXML,xml:rec}
encompasses a set of specifications with different semantics but a
common syntactic structure; XML {\em documents} must have a single
root element and paired tags, with attributes, which can be
nested. Thus, all XML documents have 
a tree structure (the so-called Document Object Model --DOM-- tree) with a single root element that
contains(encapsulates) all the contents of the document. Optionally, the syntax or semantics of elements and
attributes may be  determined by a Document Type Definition
(DTD) or XSchema (equivalent concept that uses XML for its
definition, \cite{XSchema}), in which case the document can be validated; however, in
most applications what is called {\em well-formed} XML is more than
enough.

\begin{figure}
\input {highlight.sty}
\noindent
\ttfamily
\hlstd{}\hlkwa{$<$?xml version=}\hlstr{"1.0"}\hlkwa{ ?$>$}\hlstd{}\hspace*{\fill}\\
\hlkwa{$<$html$>$}\hlstd{\hspace*{\fill}\\
}\hlstd{\ \ }\hlstd{}\hlkwa{$<$head$>$}\hlstd{\hspace*{\fill}\\
}\hlstd{\ \ \ \ }\hlstd{}\hlkwa{$<$title$>$}\hlstd{Test page}\hlkwa{$<$/title$>$}\hlstd{\hspace*{\fill}\\
}\hlstd{\ \ }\hlstd{}\hlkwa{$<$/head$>$}\hlstd{\hspace*{\fill}\\
}\hlstd{\ \ }\hlstd{}\hlkwa{$<$body$>$}\hlstd{\hspace*{\fill}\\
}\hlstd{\ \ \ \ }\hlstd{}\hlkwa{$<$h1$>$}\hlstd{Test page}\hlkwa{$<$/h1$>$}\hlstd{\hspace*{\fill}\\
}\hlstd{\ \ \ \ }\hlstd{}\hlkwa{$<$h2$>$}\hlstd{First test}\hlkwa{$<$/h2$>$}\hlstd{\hspace*{\fill}\\
}\hlstd{\ \ \ \ }\hlstd{}\hlkwa{$<$p$>$}\hlstd{Some stuff}\hlkwa{$<$br /$>$}\hlstd{\hspace*{\fill}\\
}\hlstd{\ \ \ \ }\hlstd{}\hlstd{Some more stuff}\hlkwa{$<$/p$>$}\hlstd{\hspace*{\fill}\\
}\hlstd{\ \ \ \ }\hlstd{}\hlkwa{$<$h2$>$}\hlstd{Second test}\hlkwa{$<$/h2$>$}\hlstd{\hspace*{\fill}\\
}\hlstd{\ \ \ \ }\hlstd{}\hlkwa{$<$h2$>$}\hlstd{That's another test}\hlkwa{$<$/h2$>$}\hlstd{\hspace*{\fill}\\
}\hlstd{\ \ }\hlstd{}\hlkwa{$<$/body$>$}\hlstd{}\hspace*{\fill}\\
\hlkwa{$<$/html$>$}\hlstd{}\hspace*{\fill}\\
\normalfont
\caption{An example simplified XHTML document. Looks like HTML, but it has an XML syntax: mainly, tags must be strictly paired.\label{fig:xhtml}}
\end{figure}
Since the IT industry has settled in different XML  dialects as information
exchange format, there is a business need for programs that transform
from one XML set of tags to another, extracting information or combining it in
many possible ways; a typical example of this transformation could be the extraction of
news headlines from a newspaper in Internet that uses XHTML (An XML version of
the Hypertext Markup Language (HTML) used in web pages, see figure \ref{fig:xhtml}). 

XSLT stylesheets (XML Stylesheet
Language for Transformations) \cite{XSLT}, also called {\em
logicsheets}, are designed for this purpose: applied to an XML
document, they produce another. There are  
other possible 
solutions: programs written in any language that work with text as
input and output,  programs using regular expressions or SAX filters
\cite{wiki:sax}, that process each tag in a XML document in a
different way, and do not need to load into memory the whole XML
document. However, they need external
languages to work, while XSLT is a part of the XML set of standards, and, in
fact, XSLT logicsheets are XML documents, which can be integrated
within an XML framework; that is why XSLT is, if not the most common,
at least a quite usual way of transforming XML documents. 

The amount of work needed for logicsheet creation is a problem that
scales quadratically with the quantity of initial and final formats. For
$n$ input and $m$ output formats, $n \times m$ transformations will be
needed\footnote{If an intermediate language is used, just $n + m$,
but this increases the complexity of the transformation and decreases
its speed.}. Considering that each conversion is a hand-written program
and the initial and final formats can vary with certain frequency, any
automation of the process means a considerable saving of effort on
the part of the programmers. 

The objective of this work is to find the XSLT logicsheet
that, from one input XML document, is able to obtain an output XML
document that contain exclusively the information that is considered
important from original XML documents. This information may be ordered
in any  possible way, possibly in an order different to the input
document. This logicsheet will be evolved using evolutionary operators
that will take into account the structure of the program and its
components. This could be considered, in a way, Genetic Programming,
since XSLT logicsheets are XML documents that have a tree structure,
but, since they have to follow grammatical conventions, it is better
to guide evolution using specific operators than allow all type of GP
operators. 

Thus, XSLT provides a general mechanism for the
association of patterns in the source XML document to
the application of format rules to these elements, but in order to simplify
the search space for the evolutionary algorithm, only three
instructions of XSLT will be used in this work: {\sf template}, which
sets which XML fragment will be included when the element in its
\textsf{match} attribute is found; {\sf
apply-templates}, which
is used to select the elements to which the transformation is going to
be applied and delegate control to the corresponding {\sf templates};  and {\sf value-of}\footnote{With {\sf text} used for easy
visualization of the final document}, which simply includes the
content of an XML document into the output file. This implies also a simplification of
the general XML-to-XML transformation problem: we will just extract
information from the original document, without adding new elements
(tags) that did not exist in the original document. In fact, this
makes the problem more similar to the creation of an {\em scraper}, or
program that extracts information from legacy websites or
documents. Thus, we intend this paper just as a proof of concept and
initial performance measurement,
whose generalization, if not straightforward, is at least possible. 
\begin{figure}[htb]
\input {highlight.sty}
\noindent
\ttfamily
\hlstd{}\hlkwa{$<$?xml version=}\hlstr{"1.0"}\hlkwa{?$>$}\hlstd{}\hspace*{\fill}\\
\hlkwa{$<$xsl:stylesheet version=}\hlstr{"1.0"}\hlkwa{ xmlns:xsl=}\hlstr{"http://www.w3.org/1999/XSL/Transform"}\hlkwa{$>$}\hlstd{\hspace*{\fill}\\
}\hlstd{\ \ }\hlstd{}\hlkwa{$<$xsl:output method=}\hlstr{"xml"}\hlkwa{ indent='yes'/$>$}\hlstd{\hspace*{\fill}\\
}\hlstd{\ \ }\hlstd{}\hlkwa{$<$xsl:template match=}\hlstr{"/"}\hlkwa{ $>$}\hlstd{\hspace*{\fill}\\
}\hlstd{\ \ \ \ }\hlstd{}\hlkwa{$<$output$>$}\hlstd{\hspace*{\fill}\\
}\hlstd{\ \ \ \ \ \ }\hlstd{}\hlkwa{$<$xsl:apply{-}templates select='html' /$>$}\hlstd{\hspace*{\fill}\\
}\hlstd{\ \ \ \ }\hlstd{}\hlkwa{$<$/output$>$}\hlstd{\hspace*{\fill}\\
}\hlstd{\ \ }\hlstd{}\hlkwa{$<$/xsl:template$>$}\hlstd{\hspace*{\fill}\\
}\hlstd{\ \ }\hlstd{}\hlkwa{$<$xsl:template match='html'$>$}\hlstd{\hspace*{\fill}\\
}\hlstd{\ \ \ \ }\hlstd{}\hlkwa{$<$xsl:apply{-}templates select='body'/$>$}\hlstd{\hspace*{\fill}\\
}\hlstd{\ \ }\hlstd{}\hlkwa{$<$/xsl:template$>$}\hlstd{\hspace*{\fill}\\
}\hlstd{\ \ }\hlstd{}\hlkwa{$<$xsl:template match='body'$>$}\hlstd{\hspace*{\fill}\\
}\hlstd{\ \ \ \ }\hlstd{}\hlkwa{$<$xsl:apply{-}templates select='h2'/$>$}\hlstd{\hspace*{\fill}\\
}\hlstd{\ \ }\hlstd{}\hlkwa{$<$/xsl:template$>$}\hlstd{\hspace*{\fill}\\
}\hlstd{\ \ }\hlstd{}\hlkwa{$<$xsl:template match='h2'$>$}\hlstd{\hspace*{\fill}\\
}\hlstd{\ \ \ \ }\hlstd{}\hlkwa{$<$line$>$}\hlstd{}\hlkwa{$<$xsl:value{-}of select='.' /$>$}\hlstd{}\hlkwa{$<$/line$>$}\hlstd{\hspace*{\fill}\\
}\hlstd{\ \ }\hlstd{}\hlkwa{$<$/xsl:template$>$}\hlstd{}\hspace*{\fill}\\
\hlkwa{$<$/xsl:stylesheet$>$}\hlstd{}\hspace*{\fill}\\
\mbox{}
\normalfont
\caption{Example XSLT logicsheet that extracts the content of {\tt h2} tags to an XML file; each content will be contained in {\tt line} tags. This example shows an structure with {\sf templates} for all elements in the path that leads to the element being extracted. \label{fig:xslt:1}}
\end{figure}
\begin{figure}[htb]
\input {highlight.sty}
\noindent
\ttfamily
\hlstd{}\hlkwa{$<$?xml version=}\hlstr{"1.0"}\hlkwa{?$>$}\hlstd{}\hspace*{\fill}\\
\hlkwa{$<$xsl:stylesheet version=}\hlstr{"1.0"}\hlkwa{ xmlns:xsl=}\hlstr{"http://www.w3.org/1999/XSL/Transform"}\hlkwa{$>$}\hlstd{\hspace*{\fill}\\
}\hlstd{\ }\hlstd{}\hlkwa{$<$xsl:output method=}\hlstr{"xml"}\hlkwa{ indent='yes'/$>$}\hlstd{\hspace*{\fill}\\
}\hlstd{\ }\hlstd{}\hlkwa{$<$xsl:template match=}\hlstr{"/"}\hlkwa{ $>$}\hlstd{\hspace*{\fill}\\
}\hlstd{\ \ }\hlstd{}\hlkwa{$<$output$>$}\hlstd{\hspace*{\fill}\\
}\hlstd{\ \ \ }\hlstd{}\hlkwa{$<$xsl:apply{-}templates select='/html/body/h2'/$>$}\hlstd{\hspace*{\fill}\\
}\hlstd{\ \ }\hlstd{}\hlkwa{$<$/output$>$}\hlstd{\hspace*{\fill}\\
}\hlstd{\ }\hlstd{}\hlkwa{$<$/xsl:template$>$}\hlstd{\hspace*{\fill}\\
\hspace*{\fill}\\
}\hlstd{\ }\hlstd{}\hlkwa{$<$xsl:template match='h2'$>$}\hlstd{\hspace*{\fill}\\
}\hlstd{\ \ \ }\hlstd{}\hlkwa{$<$line$>$}\hlstd{}\hlkwa{$<$xsl:value{-}of select='.' /$>$}\hlstd{}\hlkwa{$<$/line$>$}\hlstd{\hspace*{\fill}\\
}\hlstd{\ }\hlstd{}\hlkwa{$<$/xsl:template$>$}\hlstd{}\hspace*{\fill}\\
\hlkwa{$<$/xsl:stylesheet$>$}\hlstd{}\hspace*{\fill}\\
\mbox{}
\normalfont
\caption{Another example XSLT logicsheet for extracting {\tt h2} tags,
in this case using an XPath expression to process just the needed nodes. \label{fig:xslt:2}}
\end{figure}

XSLT stylesheets combines XSLT commands with embedded XPath
\cite{XPath} expressions to map XML documents into others. For
instance, to extract all {\sf H2} elements in the XHTML example shown
in figure \ref{fig:xhtml} both XSLT logicsheets shown in figures
\ref{fig:xslt:1} and \ref{fig:xslt:2} would be valid, but
the second one is simpler, making use of a single XPath expression,
while the other one would obtain the same result using only XSLT templates. In addition, XPath provides a way to select groups of
elements ({\em node-sets}) and to filter them by using predicates allowing,
for instance, to select the element that occupies a certain position
within a node-set.

Previously, we published the initial XSLT evolution experiments
\cite{zorzano-2007}, testing different document structures and
operators. In this paper we will try to improve on those results,
choosing XSLT stylesheet structure and operators so that convergence
to solution is assured. We will try also to examine the influence of
the different operator rates on the result.

The rest of the paper is structured as follows: the state of the art
is presented in section \ref{sec:soa}. Section \ref{sec:methodology} describes the
solution presented in this work. Experiments are described in  section
\ref{sec:exp}, with the automatic generation of XSLT stylesheets for two examples 
and finally the conclusions and possible lines of future work are
presented in section \ref{sec:conc}.

\section{State of the art}
\label{sec:soa}

So far, very few papers about applying genetic programming techniques to
the 
automatic generation of XSLT logicsheets have been published; one of these, by
Scott Martens \cite{martens00},  presents a technique to find XSLT stylesheets
that transform a XML file into HTML  by using genetic
programming. Martens works on simple XML documents, like the ones shown
in its article, and uses the UNIX diff function as the basis for its
fitness function. He concludes that genetic programming is useful to
obtain solutions to simple examples of the problem, but it needs
unreasonable execution times for complex examples and might not be a
suitable method to solve this kind of problems. However, computing has
changed a lot in the latest seven years, and the time for doing it is
probably now, as we attempt to prove in this paper.

Unaware of this effort, and coming from a completely different field,
Schmidt and Waltermann \cite{schmid04} approached the problem taking
into account that XSLT is a functional language, and using functional
language program generation techniques on it, in what they call {\em
inductive synthesis}. First they create a non-recursive program, and
then, by identifying recurrent parts, convert it into a recursive
program; this is a generalization of the technique used to generate
programs in other programming languages such as LISP
\cite{blisp,summers}, and used thoroughly since the eighties
\cite{Biermann:83}. 

A few other authors have approached the general problem of generating XML
document transformations knowing the original and target structure of
the documents, as represented by its DTD: Leinonen et
al. \cite{Leinonen,Kuikka02} have proposed semi-automatic generation
of transformations for XML documents; user input is needed to define the label
association. There are also freeware programs that perform transformations on
documents from a XSchema to another one. However, they must know both
XSchemata in advance, and are not able to accomplish general
transformations on well formed XML documents from examples. 

The automatic generation of XSLT logicsheets is also a super-set of 
the problem of generating 
{\em wrappers}, that is, programs that extract information from
websites, such as the one described by Ben Miled et al. in
\cite{benmiled:wrappers}. In fact, HTML is similar in structure to XML
(and can actually be XML in the shape of XHTML), but these programs do
not generate new data (new tags), but only extract information already
existing in web sites. This is what applications such as X-Fetch Wrapper,
developed by Republica\footnote{This company no longer exists, and the
product seems to have been discontinued}, do. The company that
marketed it claims that it is able to perform transformation between
any two XML formats from 
examples. Anyway, it is not so clear that transformations are that
straightforward:  according to a
white paper found at their website, it uses a document transformation language.

\section{Methodology}
\label{sec:methodology}

XSLT stylesheets have been inserted into tree structures, making them
evolve using variation operators. Every XSLT stylesheet is evaluated
using a fitness function that is related to the difference between
generated XML and output XML associated to the example. The solution
has been programmed using JEO \cite{jeo-gecco2002}, an evolutionary
algorithm library developed at University of Granada as part of the
DREAM project \cite{LNCS2439:ID197:pp665}, which is available at
\url{http://www.dr-ea-m.org} together with the rest of the
project. All source code for the programs used to run the experiments
is available from 
\url{https://forja.rediris.es/websvn/wsvn/geneura/GeneradorXSLT/} under
an open source licence.

The generated XML documents are encapsulated within an XML tag whose
name equals the root element from the input XML; each line uses also
the tag {\sf line}, so that we can distinguish easily between intended
and unintended (generated by default templates, for instance) output
lines. Next, structures used for evolution and operators applied to them are
described. These operators work on data structures and XPath queries
within them. 

The search space over possible stylesheets is exceedingly large. In
addition, language grammar must be considered in order to avoid
syntactically wrong stylesheet generation. Due to this,
transformations are applied to predetermined stylesheet structures
which have been selected, which will be described next, along with the
operators that will be applied to them.

\subsection {Type 1 structure}
\begin{figure}[htb]
\input {highlight.sty}
\noindent
\ttfamily
\hlstd{}\hlkwa{$<$xsl:template match=}\hlstr{"/"}\hlkwa{ $>$}\hlstd{\hspace*{\fill}\\
}\hlstd{\ \ \ \ }\hlstd{}\hlkwa{$<$xsl:apply{-}templates select=}\hlstr{"/book"}\hlkwa{/$>$}\hlstd{}\hspace*{\fill}\\
\hlkwa{$<$/xsl:template$>$}\hlstd{}\hspace*{\fill}\\
\hlkwa{$<$xsl:template match=}\hlstr{"book"}\hlkwa{$>$}\hlstd{\hspace*{\fill}\\
}\hlstd{\ \ \ \ }\hlstd{}\hlkwa{$<$xsl:apply{-}templates select=}\hlstr{"chapter{[}2{]}"}\hlkwa{/$>$}\hlstd{\hspace*{\fill}\\
}\hlstd{\ \ \ \ }\hlstd{}\hlkwa{$<$xsl:apply{-}templates select=}\hlstr{"chapter{[}3{]}/para{[}5{]}"}\hlkwa{/$>$}\hlstd{\hspace*{\fill}\\
}\hlstd{\ \ \ \ }\hlstd{}\hlkwa{$<$xsl:apply{-}templates select=}\hlstr{"chapter{[}2{]}//line"}\hlkwa{/$>$}\hlstd{}\hspace*{\fill}\\
\hlkwa{$<$/xsl:template$>$}\hlstd{}\hspace*{\fill}\\
\hlkwa{$<$xsl:template match=}\hlstr{"title"}\hlkwa{$>$}\hlstd{\hspace*{\fill}\\
}\hlstd{\ \ \ \ }\hlstd{}\hlkwa{$<$line$>$}\hlstd{}\hlkwa{$<$xsl:value{-}of select=}\hlstr{"."}\hlkwa{/$>$}\hlstd{}\hlkwa{$<$/line$>$}\hlstd{}\hspace*{\fill}\\
\hlkwa{$<$/xsl:template$>$}\hlstd{}\hspace*{\fill}\\
\mbox{}
\normalfont
\caption{Example of XSLT stylesheet of type 1.\label{fig:xslt:type1}}
\end{figure}
An example of this structure is shown in figure \ref{fig:xslt:type1}.
\begin{itemize}
\item The XSLT logicsheet will have three levels of depth. First level
is the root element $<${\sf xsl:stylesheet}$>$ which is common to all XSLT
stylesheets.  
\item An undetermined quantity of $<${\sf xsl:template match=...}$>$ instructions hangs from the root element.
\item The value of {\sf match} attribute for the first template that hangs
off the root will be ``/''. This template and its content never will
be modified by applying evolution operators. The only instruction inside
this element will be {\sf apply-templates}, that will have a {\sf select}
attribute whose value will be a ``/''  followed by the root
element name, so that the rest of templates included in the stylesheet
will be processed. 
\item The values of the {\sf match} attributes for the rest of the templates
       will be simply tag names of the input XML. Every
      value will have an undetermined number of children, that will be
     \textsf{ apply-template} or\textsf{ value-of} instructions. These instructions will
      have \textsf{select} attributes, whose values will be relative XPaths, built over the template path. Those routes would include every possible XPath clause. {\sf value-of} will be used instead
      of {\sf apply-templates} when the XPath is self ({\tt .}).
\end{itemize}

This kind of structure is quite unconstrained, and relies heavily in the
use of default templates. If an element is not matched, the default
template, which includes the text inside the element, is applied. For
the example shown in figure \ref{fig:xslt:type1}, default templates will
be used for the {\sf para} and {\sf chapter} element, for instance.

\subsection{Type 2 structure}
\begin{figure}[htb]
\input {highlight.sty}
\noindent
\ttfamily
\hlstd{}\hlkwa{$<$xsl:template match=}\hlstr{"/"}\hlkwa{ $>$}\hlstd{\hspace*{\fill}\\
}\hlstd{\ \ \ \ }\hlstd{}\hlkwa{$<$xsl:apply{-}templates select=}\hlstr{"/book"}\hlkwa{/$>$}\hlstd{\hspace*{\fill}\\
}\hlstd{\ \ \ \ }\hlstd{}\hlkwa{$<$xsl:apply{-}templates select=}\hlstr{"/book/title"}\hlkwa{/$>$}\hlstd{}\hspace*{\fill}\\
\hlkwa{$<$/xsl:template$>$}\hlstd{}\hspace*{\fill}\\
\hlkwa{$<$xsl:template match=}\hlstr{"/book"}\hlkwa{$>$}\hlstd{\hspace*{\fill}\\
}\hlstd{\ \ \ \ }\hlstd{}\hlkwa{$<$line$>$}\hlstd{}\hlkwa{$<$xsl:value{-}of select=}\hlstr{"chapter{[}2{]}"}\hlkwa{/$>$}\hlstd{}\hlkwa{$<$/line$>$}\hlstd{\hspace*{\fill}\\
}\hlstd{\ \ \ \ }\hlstd{}\hlkwa{$<$line$>$}\hlstd{}\hlkwa{$<$xsl:value{-}of select=}\hlstr{"chapter{[}3{]}/para{[}5{]}"}\hlkwa{/$>$}\hlstd{}\hlkwa{$<$/line$>$}\hlstd{\hspace*{\fill}\\
}\hlstd{\ \ \ \ }\hlstd{}\hlkwa{$<$line$>$}\hlstd{}\hlkwa{$<$xsl:value{-}of select=}\hlstr{"chapter{[}2{]}//line"}\hlkwa{/$>$}\hlstd{}\hlkwa{$<$/line$>$}\hlstd{}\hspace*{\fill}\\
\hlkwa{$<$/xsl:template$>$}\hlstd{}\hspace*{\fill}\\
\hlkwa{$<$xsl:template match=}\hlstr{"/book/title"}\hlkwa{$>$}\hlstd{\hspace*{\fill}\\
}\hlstd{\ \ \ \ }\hlstd{}\hlkwa{$<$line$>$}\hlstd{}\hlkwa{$<$xsl:value{-}of select=}\hlstr{"."}\hlkwa{/$>$}\hlstd{}\hlkwa{$<$/line$>$}\hlstd{}\hspace*{\fill}\\
\hlkwa{$<$/xsl:template$>$}\hlstd{}\hspace*{\fill}\\
\hspace*{\fill}\\
\mbox{}
\normalfont
\caption{Example of XSLT stylesheet of type 2.\label{fig:xslt:type2}}
\end{figure}
An example of this structure is shown in figure \ref{fig:xslt:type2}. The main differences with the first one are:
\begin{itemize}

\item The value of the {\sf match} attribute for the first template that
hangs off the root will be ``/'' too, but, in this case it will have
an indeterminate number of children, that will be all {\sf apply-templates}
instructions, whose values  for the {\sf select} attribute will be absolute XPaths
in the input XML, that will include only single slash-separated tag
names. 

\item The values for the {\sf match} attributes for the other templates that
      hang from the XML root
will be the same values that had the {\sf select} attributes of the
{\sf apply-templates}  in the first template. Therefore, there will be
      as many {\sf template} instructions as
      the number of {\sf apply-templates} in it, and they will be in the same order.

\item Every template of the previous section will have an undetermined
number of children, and all of them will be {\sf value-of} instructions,
where the value for the {\sf select} attribute will be XPath routes relative
to the XPath absolute route of the father template. These routes would
include every mechanisms of XPath that the designed operators  allow. 

\item If the absolute route of a template has a maximum depth level
inside the XML structure, its only \textsf{value-of} child will 
      {\sf select} the self element: ``.''. 

\end{itemize}

This type of structure is more heavily constrained than Type 1; search
is thus easier, since less stylesheets are generated; being more
constrained, however, mutation and crossover are much more disruptive,
and has a rougher landscape than before. 

\subsection{Genetic operators}

The operators may be classified in two different types: the first one
consists in operators that are common to the two structures and whose
assignment is to modify the XPath routes that contains the attributes of
the XSLT instructions (specially apply-template and
value-of). Operators in the second group are used to modify the XSLT 
tree structure and take different shape in each of them (so that the
structure is kept). In order to ensure the 
existence of the elements (tags) added to the XPath expressions and
XSLT instruction attributes, every time one of them is needed it is
randomly 
selected from the input file. 

The common operators are:

\begin{itemize}
\item {\sf XSLTreeMutatorXPath(Add$|$Mutate$|$Remove)Filter}: Adds, changes
      number, or removes a cardinal
      filter to any of the XPath tags that allow it. For example:\\
      \texttt{/book/chapter $\rightarrow $ /book/chapter[4]} \\
      \texttt{/book/chapter[2] $\rightarrow $ /book/chapter[4]} \\
\texttt{/book/chapter[2] $\rightarrow $ /book/chapter}\\

\item {\sf XSLTTreeMutatorXPathAddBranch}: Adds a new tag to an XPath, chosen randomly from the existing XPaths, observing the hierarchy
      of the input XML file tree: {\tt /book/chapter $\rightarrow $
      /book/chapter/title }

\item {\sf XSLTTreeMutatorXPathSetSelf}: Replaces the deepest node tag of a XPath route by the self node.

\item {\sf XSLTTreeMutatorXPathSetDescendant}: Removes one of the
      intermediate tags from a XPath route, remaining a Descendant type
      node: \texttt{/book/chapter/title $\rightarrow$ /book//title}.

\item {\sf XSLTTreeMutatorXPathRemoveBranch}: Removes the deepest
  element tag of a XPath route, ascending a level in the XML tree. For
  example: {\tt /book/chapter/title $\rightarrow $ /book/chapter}.

\end{itemize}

Other operators change the DOM structure of the XSLT logicsheet,
although not all of them can be applied to all XSLT structural types: 

\begin{itemize}

\item {\sf XSLTTreeCrossoverTemplate}: Swaps template instructions
      sub-trees between the two {\em parents}. This is the only
      crossover-like operator.

\item {\sf XSLTTreeMutator(Add$|$Mutate$|$Remove)Template}: Inserts, changes
      or removes a template. Insertion is performed on the root
      element matching a random element. The choice of
      this random element gives higher priority to the less deeper
      tags. The position of the new template inside the tree will be
      randomly selected, and its content will be
      \texttt{apply-templates} or \texttt{value-of} tags with the select
      attribute containing XPath routes relatives to the parent template
      XPath route randomly generated using the XPath operators. Change
      operates on a random node, generating a new sub-tree; and removal
      also eliminates a random template (if there are more than two).
 
 \item {\sf XSLTTree(Add$|$Remove)Apply}: It adds or removes an
   {\texttt xsl:value-of} statement to a randomly
selected template present in the tree. The position of the new leaf
inside the sub-tree that matches the template also will be randomly
selected. The new element is randomly generated from the route that
contains its parent template instruction. The \textsf{-Remove} operator
      also deletes the template node if the removed child was the last
      remaining one, but it is not applied if there is a single template left.

 \item {\sf XSLTreeMutateApply(1$|$2)}: Changes a randomly selected child
      (1) or creates a relative XPath from the one that contains the
      father {\sf xsl:template} and the XPath of the leaf that we are going
      to modify (2).
 
\item {\sf XSLTreeSetTemplateNull}: It chooses a sub-tree template from
the XSLT tree and replaces its content by a single instruction
{\sf $<$xsl:value-of select=''.''$>$}.  

\end{itemize}

\subsection{Fitness function}

Fitness is related to the difference between the desired and the
obtained output, but it has been also designed so that evolution is
helped. Instead of using a single aggregative function, as we did in
previous papers \cite{zorzano-2007}, fitness is now a vector that
includes the number of deletions and additions needed to obtain the
target output from the obtained output, and the resulting XSLT
stylesheet length. The XSLT stylesheet is correct only if the number
of deletions and additions is 0; and minimizing length helps
removing useless statements from it. So, fitness is minimized by
comparing individuals as follows: An
individual is considered better than another\begin{itemize}
\item if the number of deletions is smaller,
\item if the number of additions is smaller, being the number of
deletions the same, or
\item if the length is smaller, being the number of
deletions/additions the same.
\end{itemize}

Separating and prioritizing the number of deletions helps guide
evolution, by trying to find first a stylesheet that includes all
elements in the target document, then eliminating unneeded elements,
while, at the same time, reducing length.

\section{Experiments and results}
\label{sec:exp}

To test the algorithm we have performed several experiments with different
XML input files and a single XML output file. The algorithm has been
executed thirty times for each input XML. Seven
different input files have been used for Type 1, leaving only the hardest
ones  for Type 2. The same input file was used for several
experiments: a RSS feed from a weblog
(\url{http://geneura.wordpress.com}) and an XHTML file. All input and
output files are available from our Subversion repository:
\url{https://forja.rediris.es/websvn/wsvn/geneura/GeneradorXSLT/xml/}.

\begin{table*}[htbp]
\begin{center}
\caption{Operator priorities (used for the roulette wheel that randomly
 selects the operator to apply) used in the
 experiments.\label{tab:priorities} }
  \begin{tabular}{lr}
   \hline
   Operator & Priority \\
	\hline
XSLTTreeMutatorXPathSetSelf & 0.10\\
XSLTTreeMutatorXPathSetDescendant & 0.24 (Only Type 1)\\
XSLTTreeMutatorXPathRemoveBranch & 0.27 (Type 2) 0.39 (Type 1)\\
XSLTTreeMutatorXPathAddBranch & 0.99 \\
XSLTTreeMutatorXPathAddFilter & 0.45 (Type 2) 0.53 (Type 1)\\
XSLTTreeMutatorXPathMutateFilter & 0.64 (Type 2) 0.69 (Type 1)\\
XSLTTreeMutatorXPathRemoveFilter & 0.83\\
   \hline 
   XSLTTreeCrossoverTemplate & 0.11 \\ 
   XSLTTreeMutatorAddTemplate & 0.2 \\
   XSLTTreeMutatorMutateTemplate & 0.10\\ 
   XSLTTreeMutatorRemoveTemplate   & 0.12\\ 
   XSLTTreeAddApply       & 0.1\\ 
   XSLTTreeMutateApply1      & 0.1\\ 
   XSLTTreeMutateApply2      & 0.14\\ 
   XSLTTreeRemoveApply       & 0.1\\ 
   XSLTTreeSetTemplateNull   & 0.03\\ 
  \end{tabular}
\end{center}
\end{table*}
The computer used to perform the experiments is a Centrino Core Duo at
1.83 GHz, 2 GB RAM, and the Java Runtime Environment 1.6.0.01. The
population was 128 for all runs, and 
the termination condition was set to 200 generations or until a
solution was found and selection was performed via a 5-Tournament;
30 experiments were run, with different random seeds, for each template
type and input document. The XML and XSLT processors were the default
ones included in the JRE standard library. The operator rates used in the experiments, which
were tuned heuristically, are
 shown in table \ref{tab:priorities}.

The new fitness function, in general, yielded better results than
previously. The algorithm was able to find an adequate XSLT stylesheet
within the pre-assigned number of generations in most cases. The
breakdown of results per input file is shown in table
\ref{tab:res:type1:found}.

\begin{table}[ht]
\caption{Number of times, out of 30 experiments, a solution is not
found within the predefined number of generations using type 1 XSLT
structure. In general, the files are in increasing complexity order,
that is why it gets harder to find a solution in the latest
examples. \label{tab:res:type1:found}}
\centering
  \begin{tabular}{lr}
Input file & Times solution not found \\
\hline
1 & 0 \\
2 & 1 \\
3 & 0 \\
4 & 0 \\
5 & 3 \\
6 & 27 \\
7 & 17 \\
\hline
\end{tabular}
\end{table}
\begin{figure*}
\centering
\includegraphics[width=12cm]{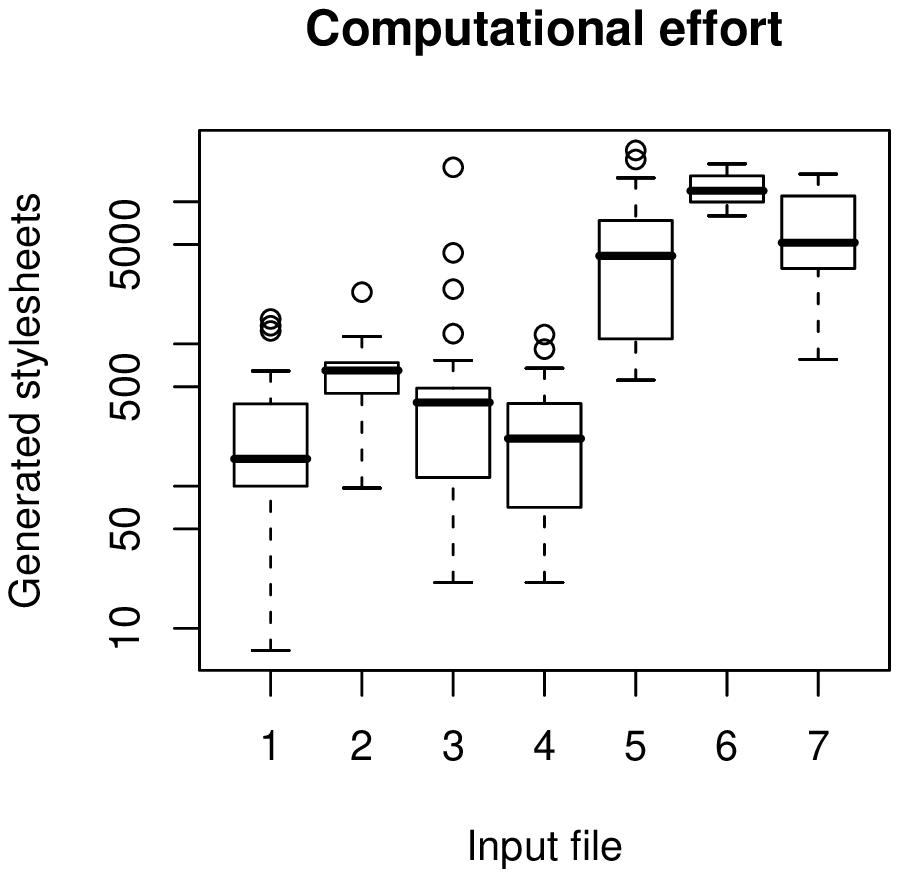}
\caption[Generados]{Logarithmic boxplot of the number of evaluations needed to
find the correct stylesheet using Type 1 structure. The difference among {\em
easy} (the first ones) and {\em difficult} (the last ones) is quite
clear; while just a few hundred of evaluations, or at most a few
thousands, are needed in files number 1 to 4, several thousands, on
average, are needed in numbers 5 and 6. Only runs when a solution was
actually found have been considered to compute averages.\label{fig:res:type1:generations}} 
\end{figure*}
\begin{figure}[htp]
\centering
\includegraphics[height=12cm]{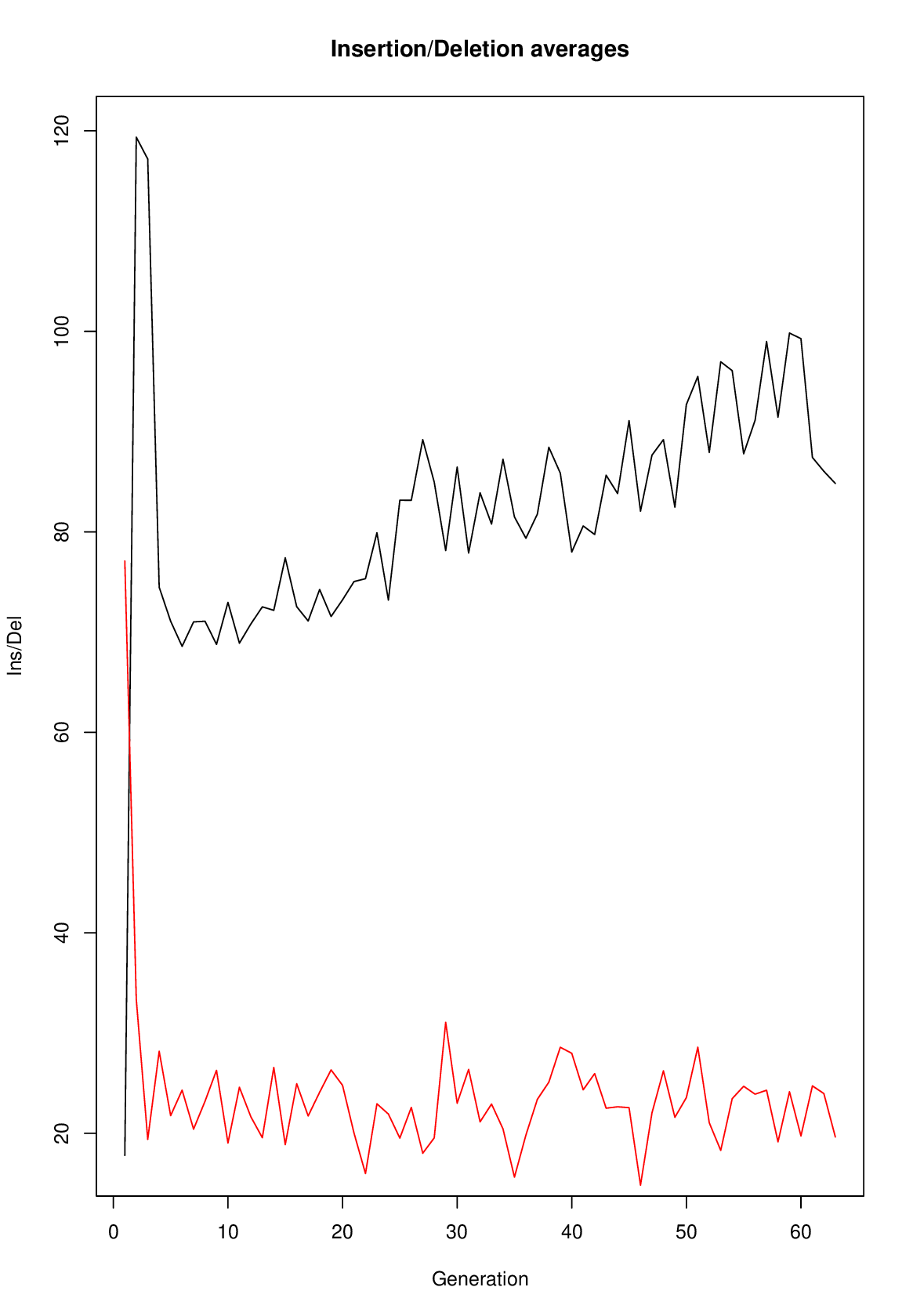}
\caption[Evolution]{Evolution of the average number of insertions
 (black, line on top) and deletions (red or light gray) for a run of file \#6 which was
 able to find a solution in around 70 generations. The number of
  deletions decreases in the first few generations, but,
 after that, it proceeds more or less randomly, exploring the search
 space until the solution is found; the number of insertions, however,
 decreases a bit after deletions' dip and then increases slowly.\label{fig:res:type1:run8}} 
\end{figure}
When a solution was found, the number of generations and time used to
find it also varies, and is shown in figure
\ref{fig:res:type1:generations}. In general, the
exploration/exploitation balance seems to be biased towards
exploration. Being such a vast and rough search space makes that, after
a few initial generations that create stylesheets with a small difference
form the target, mutations are the main operator at work, as is shown in
figure \ref{fig:res:type1:run8}.

This last figure also shows a feature of this type of evolution: every
change has a big influence on fitness, since the introduction of a
single statement can add several (dozens) lines to output. There is no
lineal relation between the number of mutations needed to reach a
solution and the number of insertions/deletions, which also means that a
single mutation might have a big influence in fitness, while several
mutations might be needed to decrease fitness by a single line. 
\begin{figure*}
\centering
\includegraphics[width=12cm]{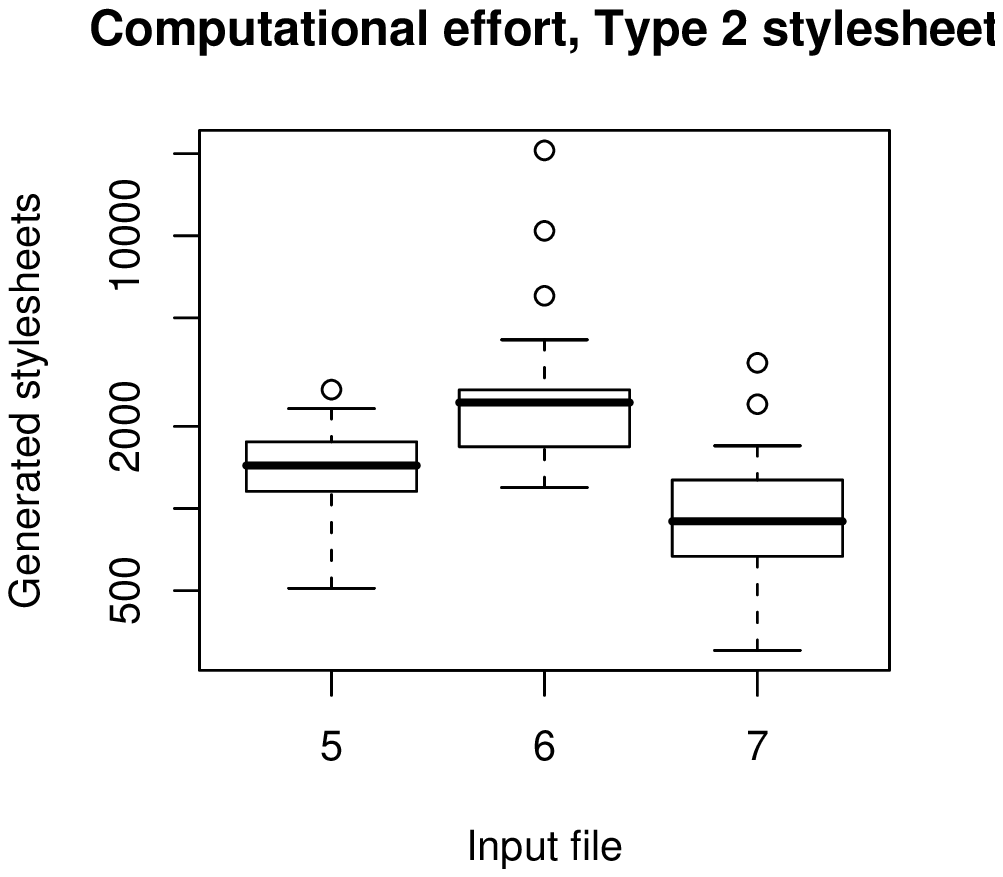}
\caption[GeneradosTipo2]{Boxplot of the number of individuals
generated to find the optimum for the Type 2 structure. File \#6
presents the maximum difficulty, needing on average around 2000
individuals. Please note that, even as finding the solution more often
than using Type 1 structure, the number of evaluations needed is smaller.\label{fig:res:type2:generations}} 
\end{figure*}

Some additional experiments have been made using type 2 structure; in
general, problems which are difficult to attack using type 1 are not so
difficult using type 2. The same number of experiments have been run
(30) for every input/output file combination, but only input files \#5, \#6
and \#7 have been used. Results are shown in figure
\ref{fig:res:type2:generations}. Once again, file \#6 presents the highest
difficulty, but using this structure raises the number of successful
experiments to 26 (out of 30); it is able to find the solution always
for the other two input files. 

In general, this structure which we have come to call Type 2 beats the
first one (Type 1) in success rate, number of generations/evaluations needed to
achieve it, and running time. The only advantage of Type 1 over Type 2
is that it has less constraints, and, in some cases, might obtain
better results; so, in general, our advice would be to try type 2
first, and if it does not yield a good result, try also type 1.

\section{Conclusions, discussion, and future work}
\label{sec:conc}

In this paper we present the results of an evolutionary algorithm
designed to search the XSLT logicsheets that is able to make a
particular transformation from a XML document to another; one of the
advantages of this application is that resulting logicsheets can be
used directly in a production environment, without the intervention of
a human operator; besides, it tackles a real-world problem found in
many organizations. Besides, it is open source software, available from
the Subversion repository
\url{https://forja.rediris.es/websvn/wsvn/geneura/GeneradorXSLT/xml/}.

In these initial experiments we have found which kind of XSLT template
structure is the most adequate for evolution, namely, one that matches
the \textsf{select} attribute in \textsf{apply-templates} with the
\textsf{match} attribute in templates, and an indeterminate number of
value-of instructions within each template; that is the one called
Type 2; this result is consistent with those found in our previous
paper \cite{zorzano-2007}. By constraining evolution
this way, we restrict the search space to a more reasonable size, and
avoid the high degree of degeneracy of the problem, with many different
structures yielding the same result, that, if combined, would result in
invalid structures. In general, we have also proved that a XSLT
logicsheet can be found just from an input/output pair of XML
documents for a wide range of examples, some of them particularly
difficult.

The experiments have shown that the search space is particularly
rough, with mutations in general leading to huge changes in
fitness. The hierarchical fitness used is probably the cause of having
a big loss of diversity at the beginning of the evolutionary search,
leading to the need of a higher level of explorations later during the
algorithm run. This problem will have to be approached via explicit
diversity-preservation mechanisms, or by using a multiobjective
evolutionary algorithm, instead of the one used now. A deeper
understanding of how different operator rates affect the result will
also help; for the time being, operator rate tuning has been very
shallow, and geared towards obtaining the result. As such, running
times and number of evaluations obtained in this paper can be used as
a baseline for future versions of the algorithm, or other algorithms
for the same problem.
 
However, there are some questions  and issues that will have to be addressed in 
future papers:

\begin{itemize}
\item Using the DTD (associated to a XML file) as a source of information
for conversions between XML documents and for restrictions of the possible
variations.
\item Adding different labels in the XSLT to allow the building of 
different kinds of documents such as HTML or WML.
\item Considering the use of advanced XML document comparison tools (i.e. XMLdiff\footnote{Available from Logilabs at \url{http://www.logilab.org/859}}).
\item Testing evolution with other kind of tools, such as a chain of SAX
      filters.
\item Obviously, testing different kinds and increasingly complex set of
      documents, and using several input and output documents at the
      same time, to test the generalization capability of the
      procedure. 
\item Using the identity transform \cite{wiki:identity:transform} as
      another frame for evolution, as an alternative to the types (which
      we have called 1 and 2) shown here. The identity transform puts
      every element found in the input document in the output document;
      elements can then be selectively eliminated via the addition of
      single statements. 
\item Tackle difficult problems from the point of view of a human
operator. In general, the XSLT stylesheets found here could have been
programmed by a knowledgeable person in around an hour, but in some
cases, input/output mapping would not be so obvious at first
sight. This will mean, in general, increase also the XSLT statements
used in the stylesheet, and also in general, adding new types of
operators. 
\end{itemize} 

\bibliographystyle{abbrv}
\bibliography{xml,xml2,geneura,xslt-evolution} 

\begin{thebibliography}{10}

\bibitem{LNCS2439:ID197:pp665}
M.~Arenas, P.~Collet, A.~Eiben, M.~Jelasity, J.~J. Merelo, B.~Paechter,
  M.~Preu\ss, and M.~Schoenauer.
\newblock A framework for distributed evolutionary algorithms.
\newblock Number 2439 in Lecture Notes in Computer Science,LNCS, pages
  665--675. Springer-Verlag, September 2002.

\bibitem{jeo-gecco2002}
M.~G. Arenas, B.~Dolin, J.-J. Merelo-Guerv\'os, P.~A. Castillo, I.~F. de~Viana,
  and M.~Schoenauer.
\newblock {JEO}: {J}ava {E}volving {O}bjects.
\newblock In W.~B. Langdon, E.~Cantú-Paz, K.~Mathias, R.~Roy, D.~Davis,
  R.~Poli, K.~Balakrishnan, V.~Honavar, G.~Rudolph, J.~Wegener, L.~Bull, M.~A.
  Potter, A.~C. Schultz, J.~F. Miller, E.~Burke, and N.~Jonoska., editors, {\em
  Poster Accepted at GECCO 2002}, page 991, 2002.

\bibitem{benmiled:wrappers}
Z.~Ben~Miled, A.~Farooq, M.~Mahoui, N.~Li, M.~Dippold, and O.~Bukhres.
\newblock A wrapper induction application with knowledge base support: A use
  case for initiation and maintenance of wrappers.
\newblock In {\em Proceedings - BIBE 2005: 5th IEEE Symposium on Bioinformatics
  and Bioengineering}, volume 2005, pages 65--72, 2005.

\bibitem{blisp}
A.~Biermann.
\newblock The inference of regular {LISP} programs from examples.
\newblock {\em IEEE Transactions on Systems, Man and Cybernetics},
  8(8):585--600, 1978.

\bibitem{Biermann:83}
A.~W. Biermann and G.~Guiho, editors.
\newblock {\em Computer Program Synthesis Methodologies}.
\newblock Reidel, Dordrecht, 1983.

\bibitem{xml:rec}
T.~Bray, J.~Paoli, C.~M. Sperberg-McQueen, and E.~Maler.
\newblock Extensible markup language ({XML}) 1.0 (second edition).
\newblock Available from {\tt http://www.w3.org/TR/2000/REC-xml-20001006},
  November 2000.

\bibitem{XSLT}
J.~Clark.
\newblock {XSL} transformations ({XSLT}), version 1.0, {W3C} recommendation 16
  november 1999.
\newblock Available from {\tt http://www.w3.org/TR/xslt.html}.

\bibitem{XPath}
J.~Clark and S.~DeRose.
\newblock {XML} path language ({XPath}), version 1.0, {W3C} recommendation 16
  november 1999.
\newblock Available from {\tt http://www.w3.org/TR/xpath}, November 1999.

\bibitem{XSchema}
D.~C. Fallside.
\newblock Xml schema part 0: Primer.
\newblock Available from \url{http://www.w3.org/TR/xmlschema-0/}.

\bibitem{xml:bible}
E.~R. Harold.
\newblock {\em {XML} {B}ible}.
\newblock {IDG} Books worldwide, 1991.

\bibitem{Kuikka02}
E.~Kuikka, P.~Leinonen, and M.~Penttonen.
\newblock Towards automating of document structure transformations.
\newblock In {\em Proceedings of the 2002 ACM Symposium on Document
  Engineering}, pages 103--110, 2002.

\bibitem{Leinonen}
P.~Leinonen.
\newblock Automating {XML} document structure transformations.
\newblock In {\em Proceedings of the 2003 ACM Symposium on Document
  Engineering}, pages 26--28, 2003.

\bibitem{martens00}
S.~Martens.
\newblock Automatic creation of {XML} document conversion scripts by genetic
  programming.
\newblock In {\em Genetic Algorithms and Genetic Programming at Stanford}, page
  269 ff., 2000.

\bibitem{LearningXML}
E.~T. Ray.
\newblock {\em Learning {XML}: creating self-describing data}.
\newblock O\'{}Reilly, January 2001.

\bibitem{schmid04}
U.~Schmid and J.~Waltermann.
\newblock Automatic synthesis of {XSL}-transformations from example documents.
\newblock In M.~Hamza, editor, {\em Artificial Intelligence and Applications
  Proceedings (IASTED International Conference on Artificial Intelligence and
  Applications (AIA 2004)}, pages 252--257, 2004.

\bibitem{summers}
P.~D. Summers.
\newblock A methodology for {LISP} program construction from examples.
\newblock {\em J. ACM}, 24(1):161--175, 1977.

\bibitem{wiki:identity:transform}
Wikipedia.
\newblock Identity transform --- {W}ikipedia{,} {T}he {F}ree {E}ncyclopedia,
  2007.
\newblock [Online; accessed 24-January-2008]:
  \url{http://en.wikipedia.org/wiki/Identity_transform}.

\bibitem{wiki:sax}
Wikipedia.
\newblock Simple {API} for {XML} --- {W}ikipedia{,} the free encyclopedia,
  2007.
\newblock [Online; accessed 21-March-2007].

\bibitem{zorzano-2007}
N.~Zorzano, D.~Merino, J.~L.~J. Laredo, J.~P. Sevilla, P.~Garcia, and J.~J.
  Merelo.
\newblock Evolving xslt stylesheets, 2007.
\newblock \url{http://xxx.arxiv.org/abs/0712.2630}.

\end{thebibliography}

\end{document}